\documentclass[letterpaper, 10 pt, conference]{ieeeconf}  

\IEEEoverridecommandlockouts                              

\makeatletter
\let\NAT@parse\undefined
\makeatother
\usepackage[numbers,sort&compress]{natbib}

\usepackage[pdftex]{graphicx}
\usepackage{amsmath}
\usepackage{amssymb}  
\usepackage{subcaption}
\usepackage{multirow}
\usepackage{array,booktabs}
\usepackage{diagbox}
\usepackage{balance}
\usepackage{xspace}
\usepackage{algorithm}
\usepackage{algpseudocode}
\usepackage{adjustbox}
\usepackage{romannum}

\usepackage[final]{hyperref}
\hypersetup{
 colorlinks=true,
 linkcolor=blue,
 filecolor=magenta,
 urlcolor=blue,
 citecolor=black
}
\usepackage{cleveref}
\crefname{figure}{Fig.}{Figs.}
\Crefname{figure}{Fig.}{Figs.}
\usepackage[font=small]{caption}
\usepackage{rpm_SIunits}
\usepackage{rpm_acronyms}
\usepackage{rpm_math}
\usepackage{rpm_misc}
\usepackage{bbm}

\usepackage{textcomp}

\usepackage{makecell}
\usepackage{stfloats} 

\usepackage{soul,color}
\usepackage{lipsum}
\usepackage{dsfont}

\usepackage{xcolor}
\usepackage{colortbl}

\usepackage{amssymb}
\usepackage{pifont}
\usepackage{tikz} 
\usepackage{subcaption}
\usepackage{caption}
\usepackage{arydshln}

\usepackage{array}
\newcolumntype{M}[1]{>{\centering\arraybackslash}m{#1}}

\title{\LARGE \bf TIDY: Thermal Infrared Image Denoising via\\Wavelet Domain Entropy and Directional Stripe Index
}

\author{Tai Hyoung Rhee${}^{1}$, Dong-Guw Lee${}^{1}$, and Ayoung Kim${}^{1*}$
\thanks{$^\dagger$This work was supported by the Institute of Information \& communications Technology Planning \& Evaluation (IITP) grant funded by the Korea government(MSIT) No.2022-0-00480, Development of Training and Inference Methods for Goal-Oriented Artificial Intelligence Agents}
\thanks{$^{1}$T. H. Rhee, D.-G. Lee, and A. Kim are with the Dept. of Mechanical Engineering, SNU, Seoul, S. Korea {\tt\small [williamrhee, donkeymouse, ayoungk]@snu.ac.kr}}%
}

\begin{document}

\maketitle \IEEEpeerreviewmaketitle
\thispagestyle{empty}
\pagestyle{empty}

\begin{abstract}

Thermal infrared (TIR) imaging has been a popular choice for field robotics due to its robust perception capability under low light visual degradation, but it suffers from severe stochastic and fixed-pattern noise that breaks downstream estimation. This noise is intensified indoors due to low thermal contrast and uniform temperature distributions, contributing to the relative lack of indoor TIR deployments. Existing TIR denoising methods exhibit a poor accuracy-efficiency tradeoff, either too slow for online deployment required in robotics or insufficiently robust to severe degradation, while typically being trained on synthetic noise. Addressing these problems, we propose TIDY, a lightweight wavelet-domain denoiser trained on real clean-noisy TIR data. By reformulating TIR denoising in the wavelet domain, TIDY explicitly disentangles noise from structural content, enabling targeted suppression with reduced spatial complexity, significantly improving inference speed over prior methods ($\sim$34Hz). TIDY introduces two new metrics, Wavelet Entropy and Wavelet Directional Stripe Index, as complementary loss terms to explicitly suppress stochastic noise and stripe artifacts. Across severe indoor corruption and zero-shot settings, TIDY improves robustness and yields consistent gains in downstream robotics tasks including thermal inertial odometry and monocular depth estimation. Code and dataset is available at: \href{https://github.com/williamrheeth/TIDY}{https://github.com/williamrheeth/TIDY}

\end{abstract}
\section{Introduction}
\label{sec:intro}

\Ac{TIR} imaging has emerged as a powerful sensing modality in robotics, capturing thermal patterns beyond the visible spectrum \cite{lee2024thermalchameleon}.
Unlike RGB cameras, TIR operates robustly in low-light and visually degraded environments (\textit{e.g.,} fog), enabling field robotics applications such as search-and-rescue \cite{cruz2021autonomous}, surveillance \cite{wong2009effective, lee2022sequential}, and hazardous inspection \cite{shin2021self}. 
Despite these advantages, most successful robotic deployments of TIR remain concentrated in outdoor environments, yet indoor robot deployment is widespread with possible lack of lighting and various visual degradation. The narrow temperature variation of indoor scenes amplifies noise in uncooled microbolometer sensors inducing stochastic fluctuations and structured \ac{FPN}, which critically degrades downstream perception and state estimation in robotic systems \cite{saputra2020deeptio, shin2022maximizing}.

This stochasticity and structured nature of \ac{TIR} noise present a challenge that classical filtering methods struggle to address. As a result, prior approaches \cite{saputra2020deeptio, zhao2020tptio, shin2022maximizing} have resorted to task-specific end-to-end deep models that implicitly absorb thermal noise, but at the cost of feature suppression, increased sample complexity, and limited online feasibility. A more intuitive alternative is to explicitly denoise the \ac{TIR} image prior to downstream processing, decoupling perception robustness from task-specific learning. This further enables direct leverage of large-scale RGB foundational models unavailable in the \ac{TIR} domain.

However, existing \ac{TIR} denoising approaches face two critical limitations. First, the lack of real clean-noisy \ac{TIR} datasets forces most methods to rely on synthetic noise generation or self-supervision, resulting in a substantial domain gap between simulated and real sensor noise. Second, state-of-the-art denoising models fail to achieve an appropriate accuracy-latency tradeoff required for robotics, as recent high-performance models \cite{thrhee-2025-icra-ws, carmichael2025trnerf} are infeasible for online execution, while lightweight models lack robustness in severely degraded indoor scenarios and exhibit weak zero-shot generalization across unseen scenes. Such limitations are particularly problematic in robotics, where labeled data is scarce and systems must operate reliably in unseen environments under online constraints. These observations motivated us to rethink \ac{TIR} denoising from a robotics perspective: prioritizing real-noise robustness, computational efficiency, and generalization capability. 

\begin{figure}
    \centering
    \includegraphics[width=1.0\linewidth]{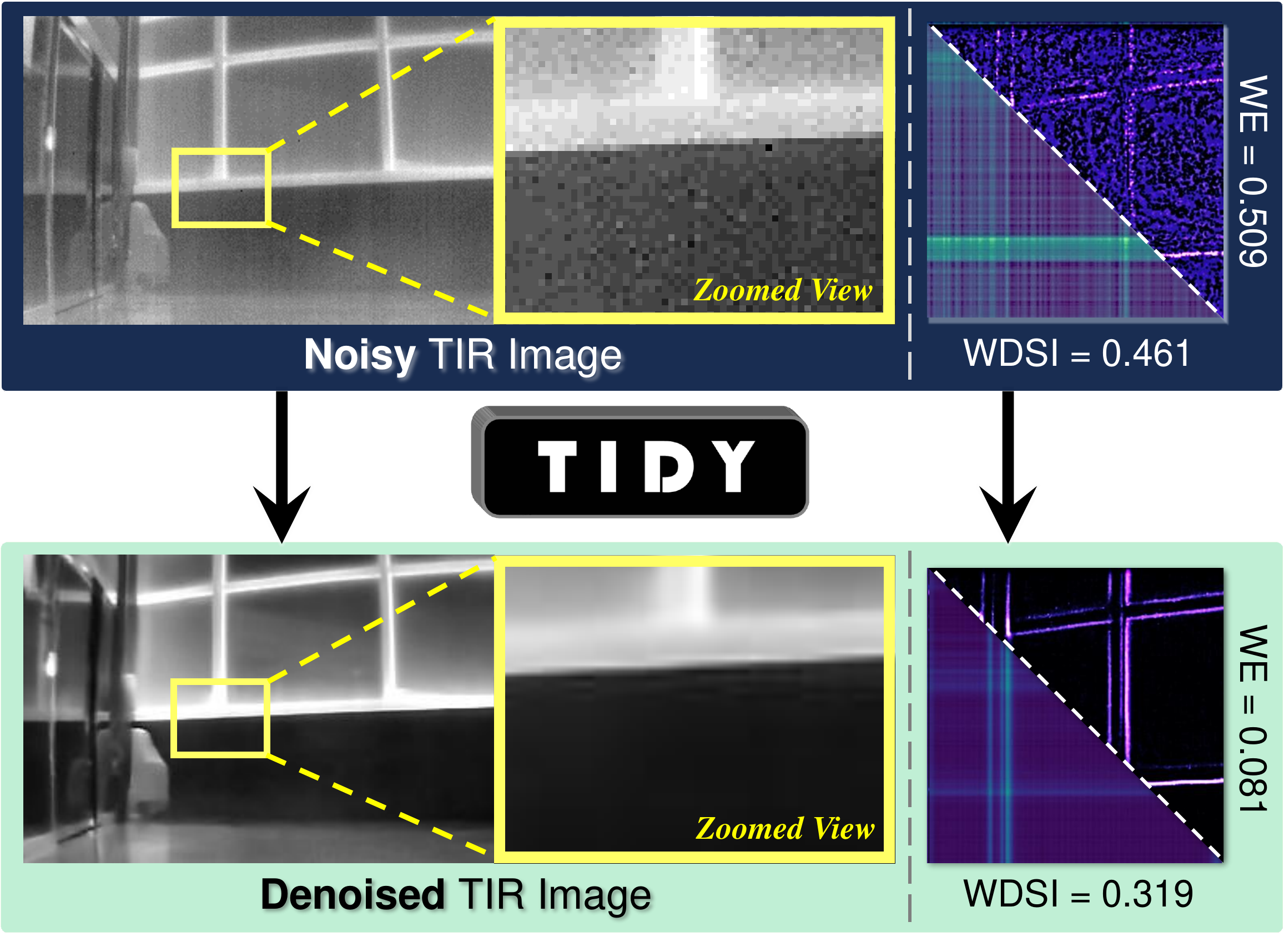}
    \caption{\textbf{TIDY} performs \ac{TIR} image denoising in the wavelet domain, reducing model complexity and effectively addressing severe noise. The proposed Wavelet Entropy (WE) and Wavelet Directional Stripe Index (WDSI) quantify random noise and \ac{FPN}, respectively, with lower values indicating less noise. The visualizations demonstrate clear suppression of stochastic speckle noise and directional stripe artifacts after denoising.}
    \label{fig:fig1}
    \vspace{-7mm}
\end{figure}

Robust denoising of real, severe thermal noise fundamentally requires explicit separation between structural content and noise components. To this end, we leverage the wavelet domain, which enables spatially localized multi-scale frequency decomposition, naturally facilitating separation of high-frequency random disturbances and stripe artifacts from scene structure. Specifically, the directional detail subbands of the \ac{DWT} inherently align with the row-/column-wise structure of \ac{FPN}, enabling explicit modeling of its directional coherence. Building on this representation, we derive statistically grounded wavelet-domain metrics that explicitly model \ac{TIR} stochastic noise and \ac{FPN}, termed \ac{WE} and \ac{WDSI}. These formulations enable direct quantification of thermal-specific noise characteristics, and can also additionally serve as a noise-aware supervisory loss, allowing explicit and robust suppression of thermal-specific noise with enhanced generalization capability.

In this paper, we introduce \textbf{TIDY}: a lightweight yet robust \underline{T}hermal \underline{I}nfrared \underline{D}enoising model leveraging Entrop\underline{y} and Directional Stripe Index in the wavelet domain. To enable training on real \ac{TIR} noise, we construct SCaN-TIR, the first real stereo clean-noisy paired \ac{TIR} dataset. Through empirical analysis of real \ac{TIR} noise, we verify that \ac{WE} and \ac{WDSI} reliably correlate with stochastic and directional \ac{FPN}, respectively, as illustrated in \figref{fig:fig1} and \figref{fig:we_wdsi}, confirming their effectiveness as thermal-specific noise indicators. Guided by these loss terms within an efficient wavelet-domain architecture that reduces spatial complexity while preserving global context via lightweight conditioning, TIDY achieves online execution and strong zero-shot generalization, particularly in severely degraded indoor scenes, translating into consistent improvements in downstream tasks such as \ac{TIO} and \ac{MDE}.



The main contributions are summarized as follows:
\begin{itemize}

    \item \textbf{Lightweight Robust \ac{TIR} Denoising Model for Robotics}: 
    We propose TIDY, a lightweight denoising network trained on real clean-noisy thermal pairs, achieving strong zero-shot generalization and online inference, with particularly robust performance in severely degraded indoor scenes.

    \item \textbf{Thermal-specific Metrics and Losses}: We introduce \ac{WE} and \ac{WDSI} as complementary metrics and losses that explicitly capture and suppress stochastic noise and directional \ac{FPN} in \ac{TIR} imagery. 

    \item \textbf{SCaN-TIR Dataset and Release}: To the best of our knowledge, SCaN-TIR Dataset is the first real stereo clean-noisy paired and aligned \ac{TIR} dataset to allow direct supervised learning of real sensor noise not synthetic approximations, with over 32.5k hardware-synced image pairs. We will publicly release both the source code and the dataset.

    
\end{itemize}

\section{related work}
\label{sec:relatedwork}

\subsection{Model-based Thermal Image Denoising}

Before the adoption of deep learning, thermal image denoising relied on spatial filtering and transform-based methods. Spatial filters such as Gaussian, median, and bilateral \cite{zhang2008multiresolution_bilateral_filtering} reduce noise locally but often blur thermal structures and degrade edges critical for robotic perception. Wavelet-based methods \cite{pathak2009the_wavelet_transform, skodras2003discrete_dwtintroduction} enable multi-scale noise suppression across frequency bands, outperforming spatial filters. However, classical approaches remain limited when handling complex noise with severe random disturbances commonly observed in TIR sensors, motivating the development of more specialized and learning-based methods.

\subsection{Data-driven Thermal Image Denoising}

Advances in deep learning have led to significant progress in thermal image denoising. Early approaches \cite{he2018single_image_based_nonuniformity-snrcnn} rely on basic \ac{CNN} models and synthetic noise patterns for supervised training, which fail to capture the complexity of real \ac{FPN}, limiting generalization. Wavelet-based networks \cite{guan2019wavelet_dnn_stripe} aim to distinguish high-frequency noise via multi-scale decomposition, but often suffer from information loss during transform operations and require careful tuning of hand-crafted components.

More recent frameworks attempt to overcome data scarcity using unsupervised or physics-inspired designs. DeepIR \cite{saragadam2021deepir} integrates domain knowledge but remains constrained by model capacity and severity of random noise of \ac{TIR} images. DestripeCycleGAN \cite{yang2024destripecyclegan} leverages simulated stripes for domain adaptation, yet introduces inconsistencies due to distribution mismatch. Methods such as TV-DIP \cite{liu2023thermal_tv-dip}, DEAL \cite{liu2025deal}, and the recent PPFN \cite{liu2025hmtir} employ priors, adversarial learning, or prompt-based conditioning to handle complex degradations, but still rely on synthetic or augmented noise. All aforementioned models exhibit poor zero-shot generalization performance. While heavier models including TIR-Diffusion \cite{thrhee-2025-icra-ws} and TRNeRF \cite{carmichael2025trnerf} yield improved results, their computational complexity renders them unsuitable for online deployment in robotic applications.  TIDY addresses these limitations with an efficient wavelet-domain design that separates noise from structure, enabling targeted removal of random and structured artifacts. Coupled with a lightweight architecture and training solely on real sensor noise, it achieves strong zero-shot generalization without synthetic data, delivering state-of-the-art online performance.  
\section{Methodology}

\begin{figure*}[t]
\vspace{1mm}  
    \centering
    \includegraphics[width=0.85\textwidth]{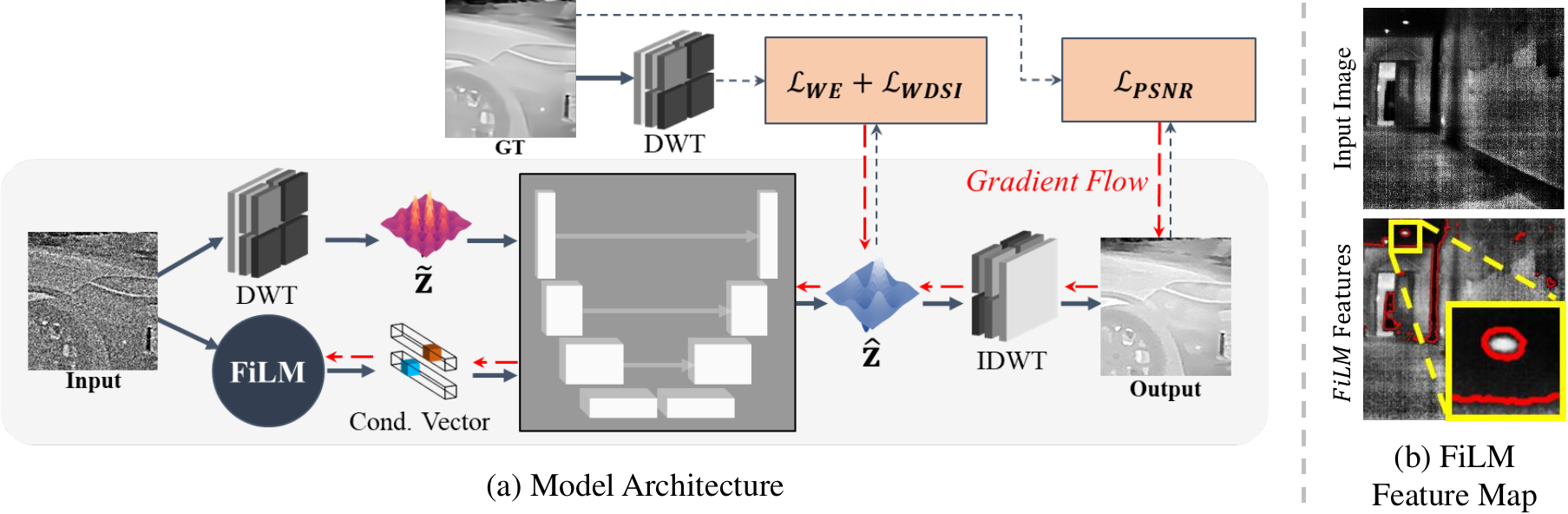}
        \caption{\textbf{(a) TIDY Model Architecture.} $\tilde{z}, \hat{z}$: wavelet coefficients of noisy, denoised image, respectively. \textbf{(b) FiLM Feature Map}: FiLM's top-10\% most strongly modulated spatial positions in the feature map in red. FiLM is trained to be strongly modulated in detailed regions prone to information loss from downsampling (\textit{e.g.,} small ceiling lamp), alleviating information loss from DWT.}
    \label{fig:model}
    \vspace{-5mm}
\end{figure*}

\subsection{Model Architecture}

TIDY's model architecture is as illustrated in \figref{fig:model}\textcolor{blue}{a}. Given an 8-bit thermal infrared image input, TIDY first applies \ac{DWT} to obtain multi-scale sub-band coefficients, where high-frequency noise and low-frequency information become spectrally separable. These coefficients are processed by a lightweight denoising backbone NAFNet \cite{chen2022nafnet} with FiLM modulation \cite{perez2018film} to preserve global context and prevent information loss from \ac{DWT}. The network predicts denoised wavelet coefficients $\hat{z}$, which are reconstructed into the image domain via \ac{IDWT}. As a plug-and-play network, TIDY only requires an image as input, and can be seamlessly attached in front of any \ac{TIR} task. Architectural details are provided below. 


\subsubsection{\ac{DWT} and \ac{IDWT}}
\label{sec:dwt_film}

The discrete wavelet transform (DWT) provides a multi-resolution representation of a signal by projecting it onto a set of scaled and translated basis functions. For a signal input $x[n]$, the \ac{DWT} can be expressed as convolution with a low-pass filter $h[\cdot]$ and a high-pass filter $g[\cdot]$ followed by dyadic downsampling, yielding the 1D approximation and detail coefficients $a[n]$ and $d[n]$, respectively \cite{pathak2009the_wavelet_transform}:

\begin{equation}
    a[n]=\sum_kx[k]\,h[2n-k]\,, \quad d[n]=\sum_kx[k]\,g[2n-k]
\end{equation}

For an image input $\mathbf{x}$, separable 2D \ac{DWT} applies this operation along rows and columns, yielding an approximation subband $cA$ and directional detail subbands $cH, cV, cD$, corresponding to horizontal, vertical, and diagonal components. This decomposition preserves spatial locality while separating structural content from high-frequency perturbations.

As illustrated in \figref{fig:model}\textcolor{blue}{a}, TIDY utilizes \ac{DWT} and \ac{IDWT} as an encoder and decoder, allowing denoising within the wavelet domain. Operating in the wavelet domain not only enhances spectral separation between noise and content, but also reduces spatial resolution, yielding computational efficiency gains.
\ac{DWT} of level $j$ leads to downsampling of the input image by a factor of $2^j$. For an input image of resolution $[H,W]$ and batch size of $B$, the \ac{DWT} alters the input from $[B,1,H,W] \rightarrow [B,4,\frac{H}{2^j},\frac{W}{2^j}]$, leading to an approximate increase in computational efficiency of $4^{j}$ as the time complexity of NAFNet for input size $[B,C,H,W]$ is $O(B\cdot C^2\cdot H\cdot W)$, with the added channels ($1\rightarrow4$) only affecting the first 1$\times$1 point-wise convolution. TIDY utilizes the Haar wavelet \cite{haar1911theorie} due to its orthogonality and compact support \cite{thrhee-2025-icra-ws}, and the \ac{DWT} encoder utilizes the empirically optimal level of $j=1$.

\begin{figure}[hb!]
\vspace{-5mm}
    \centering
    \includegraphics[width=0.9\columnwidth]{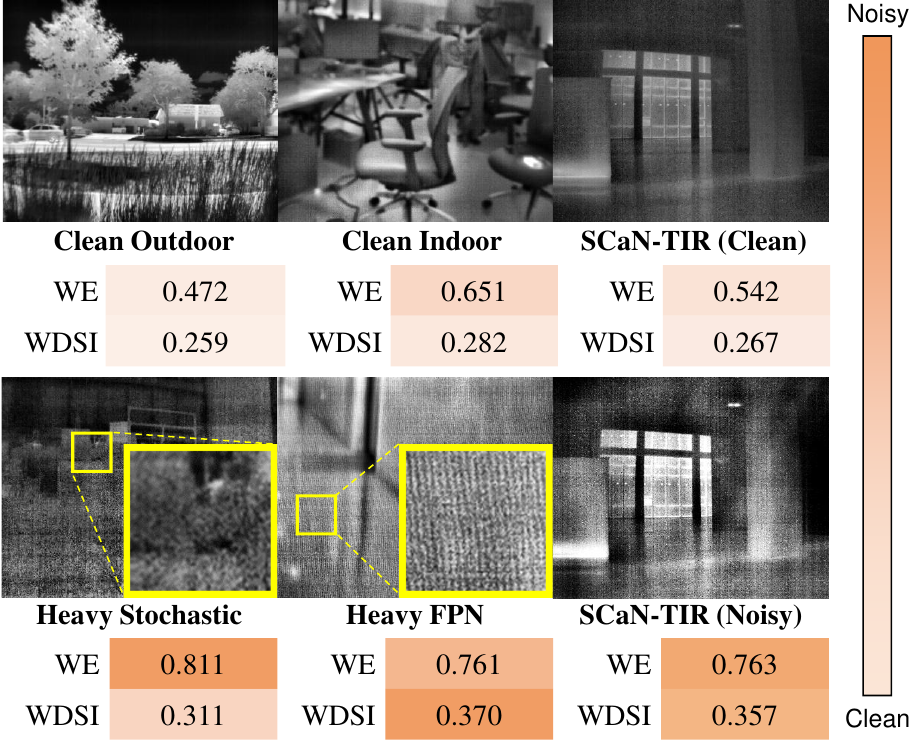}
        \caption{Comparison of WE and WDSI of various datasets \cite{dai2021multispectraldataset, carmichael2025nsavp, li2022odombeyondvision}. Clean images yield lower \ac{WE} and \ac{WDSI}, whereas heavy stochastic noise markedly elevates \ac{WE} and directional \ac{FPN} increases \ac{WDSI}, validating that WE primarily captures random noise intensity and WDSI responds to directional stripe artifacts.}
    \label{fig:we_wdsi}
    \vspace{-1mm}
\end{figure}

\subsubsection{FiLM Conditioning}

However, the efficiency gain from \ac{DWT} comes at the cost of information and global context, a fundamental shortcoming of previous wavelet-based methods. To mitigate this informational loss, we integrate a lightweight FiLM \cite{perez2018film} conditioner which injects global image-level information back into the wavelet feature space. Specifically, a global representation is extracted from the original noisy input and mapped to per-channel scale $\gamma$ and shift parameter $\beta$, which modulate intermediate wavelet features. After the initial convolution compresses the four wavelet subbands to the network’s base width, FiLM adaptively recalibrates each feature channel as
\begin{equation}
    \mathbf{x} \leftarrow \gamma \odot \mathbf{x} + \beta
\end{equation}
thereby restoring contextual awareness and selectively amplifying structurally meaningful responses (\textit{e.g.,} edges and fine details) that may be attenuated during wavelet decomposition.
FiLM’s projector is trained end-to-end alongside NAFNet and seamlessly improves fine detail reconstruction without adding significant computational overhead. \figref{fig:model}\textcolor{blue}{b} shows the top-10\% most strongly FiLM-modulated feature locations in red, which is concentrated on edges and small features like the light sources on the ceiling, preventing such features from being excluded via \ac{DWT}.  


\subsection{Loss Targeting Thermal-specific Noise}


\begin{figure}[t]
    \centering
    \vspace{1mm}
    \includegraphics[width=0.98\columnwidth]{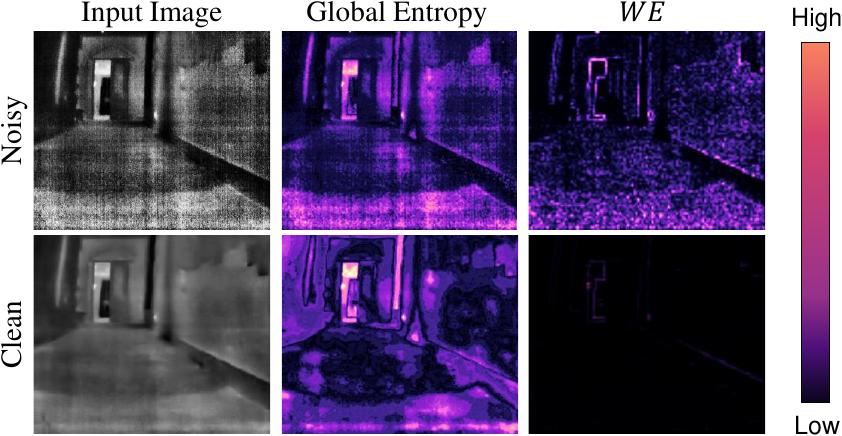}
        \caption{Comparison between global Shannon entropy and WE. WE exhibits superior sensitivity to thermal noise, yielding strong responses in noisy regions and minimal activation in clean scenes.}
    \label{fig:we_entropy}
    \vspace{-6mm}
\end{figure}

As mentioned in \secref{sec:intro}, uncooled microbolometer \ac{TIR} camera is typically corrupted by (i) Gaussian noise and (ii) \ac{FPN}. We target these two noise types separately via Wavelet Entropy (WE) and Wavelet Directional Stripe Index (WDSI), respectively. Both values also function as quantitative metrics of noise, as shown in \figref{fig:we_wdsi}, where clean images exhibit lower \ac{WE} and \ac{WDSI}, while heavy stochastic noise and directional \ac{FPN} significantly increases \ac{WE} and \ac{WDSI}, respectively.

\subsubsection{Wavelet Entropy (WE)}

\ac{WE} loss is designed to suppress stochastic Gaussian noise. Let the observed \ac{TIR} image $I$ be corrupted with additive zero-mean Gaussian noise $G \sim \mathcal{N}(0, \sigma^2)$. 
The differential entropy of the noise is:
\begin{align}
    H(G)
    &= \frac{1}{2}\log (2\pi e\sigma^2)
\end{align}
where monotonicity in variance $\sigma^2$ is immediate:
\begin{equation}
    \frac{d}{d(\sigma^2)}H(G)=\frac{1}{2\sigma^2}>0
\end{equation}
\noindent implying that entropy increases strictly with variance. 

However, directly applying an entropy penalty in the pixel domain conflates noise randomness with natural image intensity variations, as visually observed in \figref{fig:we_entropy}. Pixel-domain global entropy responds strongly to structural content and edges, even in clean images.
To ensure that entropy targets noise not image content, we evaluate entropy in the wavelet detail domain. Applying a multi-level 2D \ac{DWT} to $I$, the resulting detail subbands $K\in {H, V, D}$ have image content largely attenuated, making entropy a direct proxy for the variance of stochastic Gaussian noise. As shown in \figref{fig:we_entropy}, wavelet entropy suppresses structured regions while amplifying stochastic noise patterns.

Specifically, for each detail subband $k\in K$, entropy is computed by constructing a normalized histogram $p_j^k=\frac{\sum{|c_j^k|}}{\sum_{k'\in K}\sum |c_j^{k'}|}$, where $c_j^k$ is the wavelet coefficients for subband $k$ at decomposition level $j$. 
The \ac{WE} loss is obtained by the normalized mean entropy across all three directional subbands and across $J$ wavelet decomposition levels, which is scaled via power transform of $\gamma$:
\begin{equation}
    \mathrm{WE} = \left(\frac{1}{J \log(3)} \sum^J_{j=1} \left[-\sum_{k\in K} p_j^k \log(p_j^k)\right] \right)^\gamma 
\end{equation}

This quantity, normalized to the range $(0,1)$, measures the randomness of the wavelet coefficient distribution, with higher values indicating more noise-like behavior. 

\subsubsection{Wavelet Directional Stripe Index}

\ac{TIR} images are also affected by structured \ac{FPN}, most notably stripe artifacts arising from row-/column-wise gain and offset non-uniformities in the sensor and readout chain. Unlike stochastic noise, stripe noise is spatially correlated and exhibits strong directional coherence, which is not adequately captured by entropy. To quantify this directional coherence, we exploit the orientation selectivity of wavelet detail subbands, where row-aligned stripes primarily affect the horizontal subband $cH_j$ and column-aligned stripes the vertical subband $cV_j$. Based on this observation, the horizontal and vertical \ac{WDSI} are defined as:
\begin{equation}
    \mathrm{WDSI}_h(j)=\frac{\sigma(\mu_{row}(cH))}{\sigma(cH)}\space,\space\space \mathrm{WDSI}_v(j)=\frac{\sigma(\mu_{col}(cV))}{\sigma(cV)}
\end{equation}

\noindent where $\mu_{row}(\cdot)$ and $\mu_{col}(\cdot)$ denote averaging across rows and columns, respectively, and $\sigma(\cdot)$ denotes the standard deviation. This metric captures the relative prominence of directional patterns by measuring the variability of mean values in each orientation, normalized by the overall coefficient variability. The net \ac{WDSI} loss is averaged across multiple wavelet decomposition levels $J$ as:
\begin{equation}
    \mathrm{WDSI}=\frac{1}{J}\sum^J_{j=1} \frac{1}{2} (\mathrm{WDSI}_h(j)+\mathrm{WDSI}_v(j))
\end{equation}

\ac{WDSI} loss targets directional non-uniformity by penalizing imbalanced statistical variations along vertical and horizontal wavelet components, thereby reducing structured stripe artifacts commonly found in the \ac{TIR} domain.

Combined with the pixel-level PSNR loss $\mathcal{L}_{\mathrm{PSNR}}$ \cite{chen2022nafnet}, the net loss is formulated as a weighted sum as below:
\begin{equation}
    \mathcal{L}_{net} = \mathcal{L}_{\mathrm{PSNR}} + \lambda_{\mathrm{WE}} \mathrm{WE} + \lambda_{\mathrm{WDSI}}\mathrm{WDSI}
\end{equation}
Empirically optimal values of $J=3$, $\lambda_{\mathrm{WE}}=0.2$ and $\lambda_{\mathrm{WDSI}}=0.4$ were utilized.

\section{experiment}
\label{sec:experiment}

\subsection{The SCaN-TIR Dataset}

\begin{figure}[t]
    \vspace{1mm}
    \centering  
    \includegraphics[width=0.96\columnwidth]{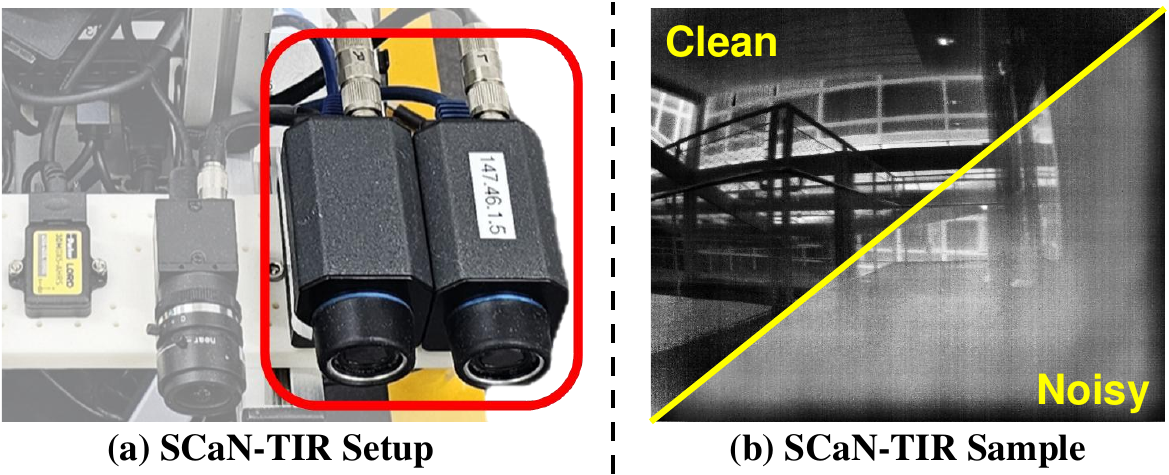}
    \caption{\textbf{SCaN-TIR.} \textbf{(a)} Setup for SCaN-TIR Dataset, with the \textcolor{red}{red} box showing the two adjacently placed A65 \ac{TIR} on an UGV. \textbf{(b)} Comparison of clean and noisy sample from SCaN-TIR Dataset.}
    \label{fig:scantir}
    \vspace{-6mm}
\end{figure}

SCaN-TIR dataset, the first ever \underline{S}tereo \underline{C}lean-\underline{a}nd-\underline{N}oisy Paired Thermal Infrared Dataset, is composed of 32.5k real paired and aligned \ac{TIR} images of dimension $640\times512$, with both indoor and outdoor sequences. As shown in \figref{fig:scantir}\textcolor{blue}{a}, two FLIR A65 IR cameras were placed adjacently, with \ac{NUC} turned off for one camera. The two camera streams were geometrically aligned by stereo rectification based on the calibrated extrinsics obtained via DiscoCal \cite{song2024discocal}. Prior to data acquisition, noise was accumulated for over 4 hours with \ac{NUC} off for the noisy camera, leading to real clean-noisy image pairs as shown in \figref{fig:scantir}\textcolor{blue}{b}. SCaN-TIR contains 9 continuous stereo sequences across diverse indoor and outdoor scenes, with each sequence recorded in a distinct real-world setting.

\begin{figure*}[hb!]
    \vspace{-3mm}
    \centering
    \includegraphics[width=0.95\linewidth]{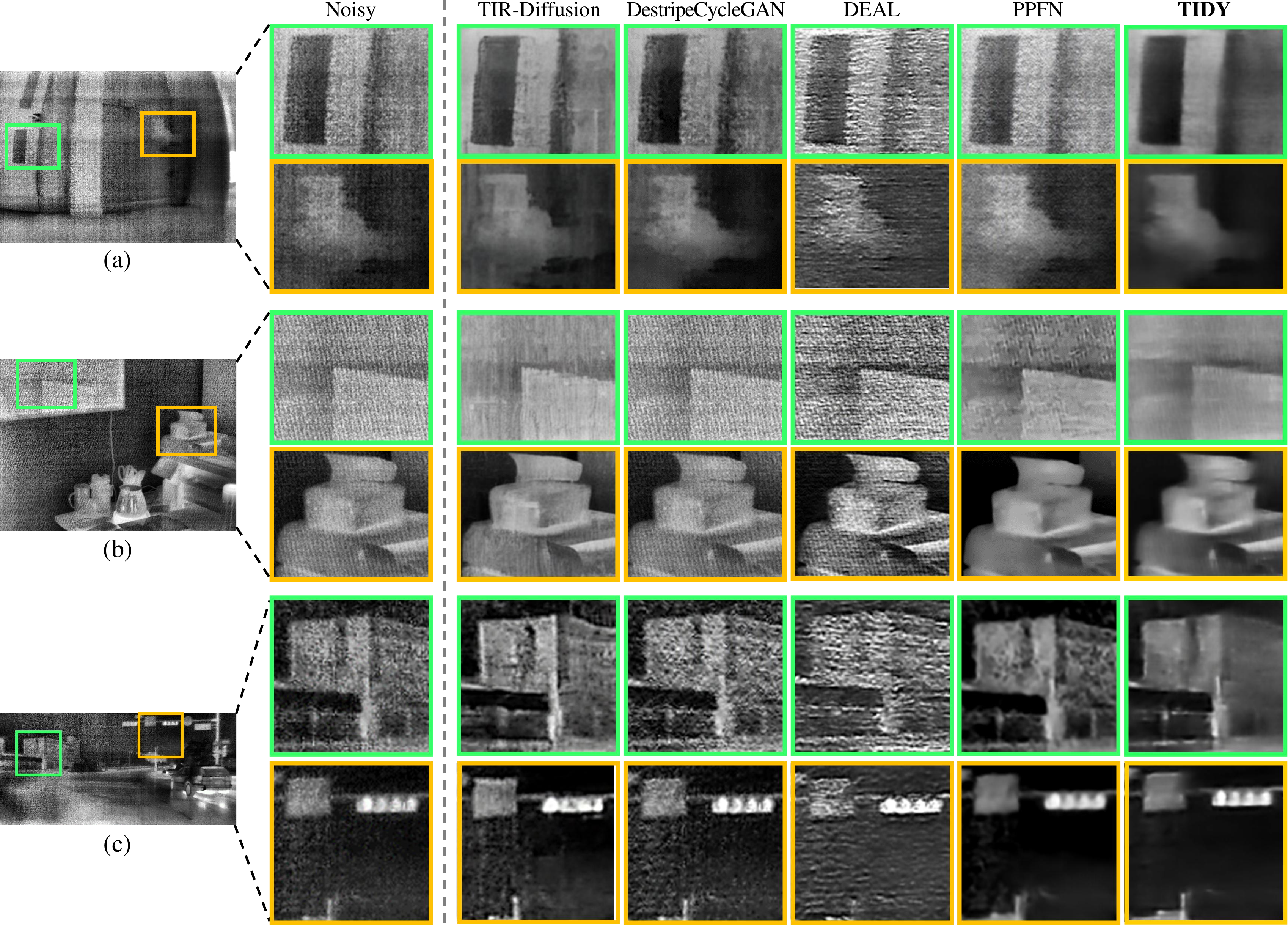}
    \vspace{-2mm}
    \caption{\textbf{Zero-shot Performance.} Denoising results on two indoor datasets and two outdoor datasets. Colored boxes denote zoomed views. TIDY produces cleaner textures and sharper structural details compared to prior methods, particularly in flat regions and fine edges.}
    \label{fig:zeroshot_results}
\end{figure*}

\subsection{Experiment Setup}

\noindent \textbf{Datasets.} For training and testing, two datasets were utilized separately: the IRE dataset \cite{liu2025twostage_irenhance}, composed of artificial additive noise, and our SCaN-TIR dataset, composed of real noise, was used for comparing TIDY with existing methods. The official train-test split was adopted for IRE, while for SCaN-TIR the \texttt{303\_floor7} sequence was designated as the test set and all remaining sequences were used for training.
Models were further experimented on zero-shot cases of existing \ac{TIR} datasets including OdomBeyondVision \cite{li2022odombeyondvision}, Multi-Spectral dataset \cite{dai2021multispectraldataset}, and MS$^2$ dataset \cite{shin2023ms2}.



\noindent \textbf{Implementation Detail.} TIDY was trained for 80k iterations with a batch size of 8, using the AdamW optimizer with an initial learning rate of 0.001 and a cosine annealing schedule applied over the first 40k iterations. All training was done on a single RTX 4090 GPU for approximately 4 hours, with a mean inference rate of 34Hz for 640$\times$512 input resolution. All experiments were conducted with the latest 8-bit tone mapping method Fieldscale \cite{gil2024fieldscale}.

\setlength\dashlinedash{2pt}
\setlength\dashlinegap{2pt}

\begin{table}[t]
\vspace{1.5mm}
\caption{Numerical Comparisons of Various Models on IRE \cite{liu2025twostage_irenhance} and our SCaN-TIR Dataset, and their operational FPS. Best results are highlighted in \textbf{bold} and second best are \underline{underlined}.}
\label{tab:results_compare}
\centering
\resizebox{1.0\columnwidth}{!}{%
\begin{tabular}{l|cc|cc|c}
\hline \hline
\multirow{2}{*}{Method} & \multicolumn{2}{c|}{IRE Dataset} & \multicolumn{2}{c|}{SCaN-TIR Dataset} & \multirow{2}{*}{FPS ↑} \\  
 & PSNR ↑ & SSIM ↑ & \multicolumn{1}{c}{PSNR ↑} & \multicolumn{1}{c|}{SSIM ↑} &  \\ \hline
Median Filter    & 28.13 & 0.6884 & 11.38 & 0.2264 & 3264 \\
Bilateral Filter & 28.73 & 0.6614 & 11.43 & 0.2464 & 193.5\\
\hdashline
U$^2$D$^2$Net & 23.63 & 0.7358 & \multicolumn{2}{c|}{\textit{N.A.}$^\dagger$} & \multicolumn{1}{c}{\textit{N.A.}$^\dagger$} \\
Liu et al. & 24.53 & 0.5335 & \multicolumn{2}{c|}{\textit{N.A.}$^\dagger$} & \multicolumn{1}{c}{\textit{N.A.}$^\dagger$} \\
DeepIR & \multicolumn{2}{c|}{\textit{N.A.}*} & 10.75 & 0.2104 & 0.05 \\
TIR-Diffusion  & 27.97 & 0.8594 & 11.45 & 0.2615 & 0.9 \\
DestripeCycleGAN  & 29.13 & 0.9007 & 11.30 & 0.2266 & \textbf{60.9} \\
DEAL  & 18.25 & 0.2093 & 9.893 & 0.0347 & 20.1 \\
PPFN & \underline{33.87} & \underline{0.9242} & \underline{12.39} & \underline{0.2829} & 10.6 \\ \hline

\textbf{TIDY} & \textbf{36.49} & \textbf{0.9547} & \textbf{14.11} & \textbf{0.3578} & \underline{34.3} \\ \hline \hline

\end{tabular}%
}

\vspace{1mm}
{\scriptsize \raggedright \parbox{\columnwidth}{*DeepIR requires sequential data, but the IRE dataset is non-sequential.}}
{\scriptsize \raggedright \parbox{\columnwidth}{$^\dagger$No open-source code provided.}}

\vspace{-6mm}
\end{table}

\subsection{Benchmark Evaluation}
\label{subsec:ire_experiment}

For quantitative benchmark evaluation, TIDY was compared against thermal denoising baselines: PPFN \cite{liu2025hmtir}, DEAL \cite{liu2025deal}, TIR-Diffusion \cite{thrhee-2025-icra-ws}, DestripeCycleGAN \cite{yang2024destripecyclegan}, DeepIR \cite{saragadam2021deepir}, U$^2$D$^2$Net \cite{ding2023u2d2net}, two-stage model by Liu et al. \cite{liu2025twostage_irenhance}, and two classical filter-based methods, median filter and bilateral filter. All learning baselines were trained identically on the train split of the chosen test datasets, following the original training protocols and hyperparameters. Sequential input of five images were utilized for DeepIR. All reported FPS values were measured on Intel i9-10900X CPU and RTX 4090 GPU with SCaN-TIR data.

As shown in \tabref{tab:results_compare}, TIDY consistently outperforms all competing methods by a large margin, while maintaining online computational efficiency. TIDY's FPS is over 2-3$\times$ higher than existing state-of-the-art models, and overall ranking second only to DestripeCycleGAN, a lightweight but legacy GAN-based model with limited representational capacity.
These results not only set a new benchmark for thermal denoising performance but also highlight TIDY’s unique ability to handle both outdoor-indoor, synthetic-real, and mild-severe noise conditions with unmatched accuracy, robustness and efficiency.

\subsection{Zero-shot Performance}

To assess the generalization ability of each model, we conducted zero-shot evaluation. TIDY trained on the SCaN-TIR dataset was employed, and pretrained weights were used for PPFN, DEAL, DestripeCycleGAN, and TIR-Diffusion.  

As shown in \figref{fig:zeroshot_results}, TIDY shows the cleanest output for both severely noisy indoor and outdoor sequences, showing clear details with minimal stripe and random noise. Flat regions, such as walls in \figref{fig:zeroshot_results}\textcolor{blue}{a} and \figref{fig:zeroshot_results}\textcolor{blue}{c}, object surfaces in \figref{fig:zeroshot_results}\textcolor{blue}{b} are effectively denoised, yielding uniform and consistent surface appearance. DestripeCycleGAN fails to recognize and retain severe noise, leading to minuscule change from the noisy image. PPFN, as the runner-up, shows decent performance in informative regions, but still fails to retain severe noise in flat regions as in \figref{fig:zeroshot_results}\textcolor{blue}{a}. DEAL introduces noisy artifacts, causing new complications and leading to downstream problems. Due to the overwhelming generalization power of diffusion, TIR-Diffusion leads to warping of original shapes as in \figref{fig:zeroshot_results}\textcolor{blue}{a}. These findings underscore TIDY’s strong zero-shot generalization, effectively handling diverse datasets and severe noise conditions without requiring dataset-specific retraining.

\begin{figure*}[!b]
    \vspace{-2mm}
    \centering
    \includegraphics[width=0.98\textwidth]{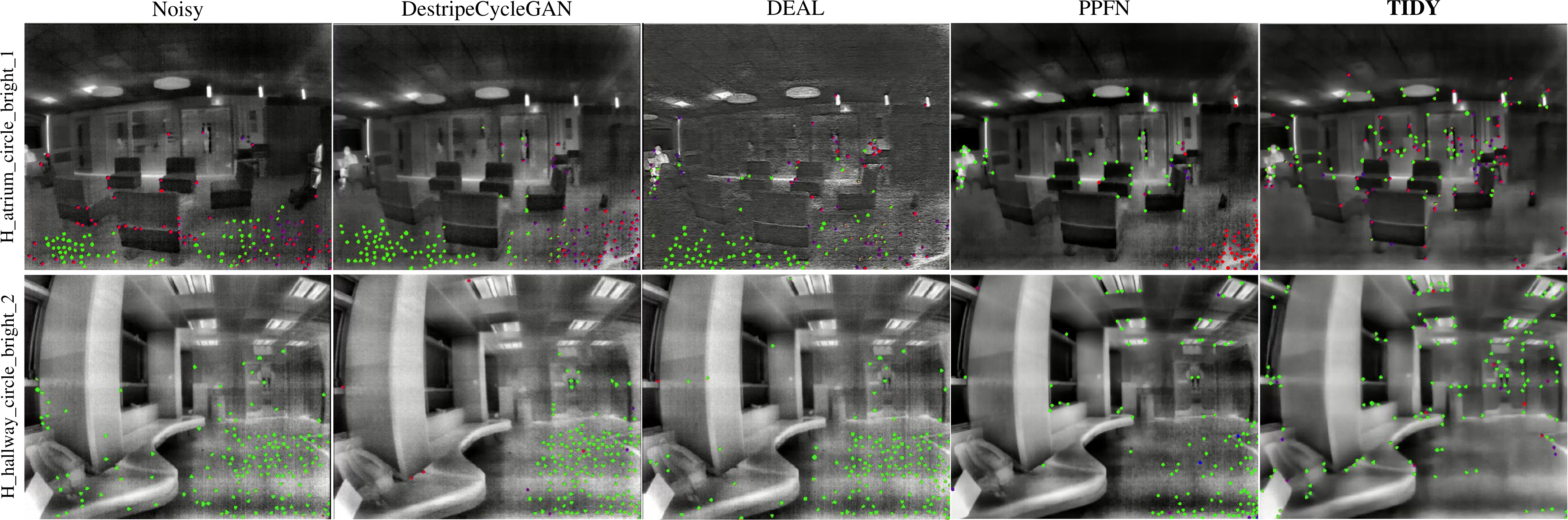}
    \caption{\textbf{TIO Performance.} Comparison of extracted features from VINS-Mono with OdomBeyondVision data. \textcolor{green}{Green}: Persistently tracked features, \textcolor{red}{Red}: New features, \textcolor{purple}{Purple}: Outliers. Features are extracted from the \ac{FPN} in the noisy images, as well as in baseline-denoised images, leading to significant errors as the \ac{FPN} is constant as the system moves. Such errors are substantially reduced via TIDY.}
    \label{fig:results_odom_features}
\end{figure*}




\subsection{Thermal Inertial Odometry}

We conduct \ac{TIO} evaluations by integrating noisy, baseline-denoised (DestripeCycleGAN, DEAL and PPFN), and TIDY-denoised thermal images into the standard VINS-Mono \cite{qin2018vinsmono} framework, averaged over 3 trials to reduce stochasticity concerns. Additionally, we compare these results against DeepTIO \cite{saputra2020deeptio}, a deep learning-based TIO method leveraging deep neural networks trained explicitly for thermal-inertial feature extraction and pose regression. TIR-Diffusion and DeepIR were not implemented due to their lack of online capability, as TIR-Diffusion offers below 1 Hz inference, and DeepIR requires an iterative procedure for a single image. TIDY-denoised VINS-Mono achieves the best performance as shown in \tabref{tab:odom_results}, resulting in the lowest \ac{ATE} in comparison to baseline denoising models and even DeepTIO for most sequences. 

\newcolumntype{C}[1]{>{\centering\arraybackslash}m{#1}}

\begin{table}[]
\vspace{2mm}
\caption{\ac{ATE} [m] of VINS-Mono with OdomBeyondVision Dataset of Baselines and TIDY, as well as Comparison with DeepTIO \cite{saputra2020deeptio}. Best results are highlighted in \textbf{bold} and second best are \underline{underlined}.}
\label{tab:odom_results}
\centering
\setlength{\tabcolsep}{2pt}
\resizebox{1.0\columnwidth}{!}{%
\begin{tabular}{l|C{1.1cm} C{1.1cm} C{1.1cm} C{1.1cm} C{1.1cm} C{1.1cm}}
\hline \hline \rule{0pt}{3.0ex}
Sequence & Noisy & \resizebox{!}{13pt}{\shortstack{Destripe\\-CycleGAN}} & DEAL & PPFN & DeepTIO & \textbf{TIDY} \\ \hline

\resizebox{!}{5pt}{H\_atrium\_circle\_bright\_2} 
& 73.94 
& 12.16
& 397.31 
& \underline{5.00} 
& \textit{N.A.$^\dagger$} 
& \textbf{4.23} \\

\resizebox{!}{5pt}{H\_atrium\_circle\_bright\_1} 
& 568.38 
& 57.91 
& 250.05 
& \underline{3.35}
& 7.24
& \textbf{2.78} \\

\resizebox{!}{5pt}{H\_vicon\_circle\_dark\_4} 
& \textit{Fail*} 
& \textit{Fail*} 
& 61.85 
& 2.88 
& \textbf{1.24} 
& \underline{1.35} \\

\resizebox{!}{5pt}{H\_hallway\_circle\_bright\_2} 
& \textit{Fail*} 
& \textit{Fail*} 
& \textit{Fail*} 
& \underline{9.19}
& 13.39
& \textbf{5.99} \\

\resizebox{!}{5pt}{G\_corridor\_linear\_bright\_13} 
& \textit{Fail*} 
& 3.78 
& 6.28 
& 1.81 
& \underline{1.32} 
& \textbf{1.20} \\

\hline \hline

\end{tabular}%
}
\\
\vspace{1mm}
{\scriptsize \raggedright \parbox{\columnwidth}{*VINS failed to initialize.}}
\\
{\scriptsize \raggedright \parbox{\columnwidth}{$^\dagger$Results of sequence not provided in paper.}}

\vspace{-6mm}
\end{table}

As depicted in \figref{fig:results_odom_features}, noisy and the baseline-denoised images primarily guide feature extraction toward fixed-pattern noise, whose spatial position remains largely invariant despite system motion, thereby leading to severe pose estimation errors and degraded odometry performance. In numerous instances with noisy and the two baseline images, VINS-Mono initialization fails due to misalignment between visual structures and IMU measurements, resulting in complete system failure. In contrast, the TIDY-denoised images enable reliable extraction of motion-consistent features, improving visual-IMU alignment and ensuring stable initialization and accurate odometry estimation. Please refer to the supplementary video for the full \ac{TIO} demonstration. 

\begin{figure}[t]
    \centering
    \vspace{1.5mm}
    \includegraphics[width=0.95\columnwidth]{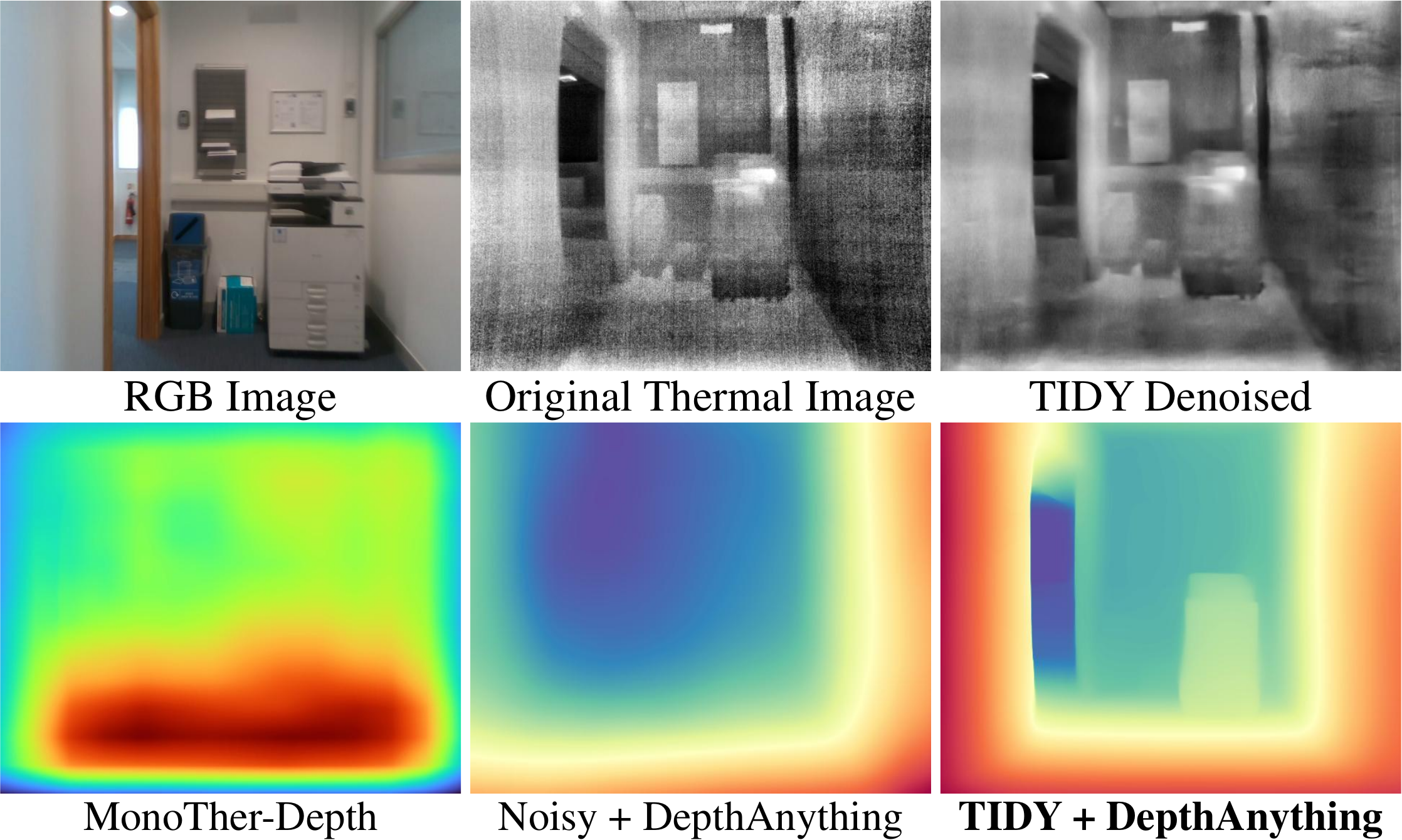}
    \caption{\textbf{MDE Performance.} Depth Anything 3 results of original noisy image from OdomBeyondVision, zero-shot denoised image using TIDY as well as comparison with thermal-specific depth estimator, MonoTher-Depth.}
    \label{fig:depthanything_result}
    \vspace{-2mm}
\end{figure}

\subsection{Monocular Depth Estimation}

To further validate TIDY’s robotics application, we utilized the pretrained Depth Anything 3 \cite{lin2025depthanything3} model to perform \ac{MDE} directly from the raw noisy and TIDY-denoised images, as well as a thermal-specific monocular depth estimator, MonoTher-Depth \cite{zuo2025monotherdepth}. As shown in \figref{fig:depthanything_result}, noisy inputs from OdomBeyondVision dataset severely degrade predicted depth maps, obscuring scene geometry and object boundaries, while MonoTher-Depth shows suboptimal results for indoor sequences with severe noise. In contrast, TIDY-denoised images yield significantly clearer and more accurate depth predictions by portraying the open path on the left and the rectangular object on the right, underscoring our model’s effectiveness in downstream thermal vision tasks. Quantitatively, when evaluated on the ViViD++ dataset in zero-shot setting, TIDY consistently reduces absolute relative error (AbsRel) while improving $\delta_1$ accuracy as shown in \tabref{tab:depth_results}, confirming systematic gains across sequences. TIDY with Depth Anything also demonstrates superior performance in \ac{MDE} for input video sequences. Please refer to the supplementary video for demonstration. 

\begin{table}[t]
\centering
\caption{\ac{MDE} results on ViViD++ \cite{lee2022vivid++} indoors sequences. Lower AbsRel is better ($\downarrow$), higher $\delta_1$ is better ($\uparrow$). Best results are highlighted in \textbf{bold}.}
\label{tab:depth_results}
\resizebox{1.0\columnwidth}{!}{%
\begin{tabular}{l|cc|cc|cc}
\hline \hline
\multirow{2}{*}{Sequence} 
& \multicolumn{2}{c|}{\rule{0pt}{2.1ex}Noisy} 
& \multicolumn{2}{c|}{\rule{0pt}{2.1ex}MonoTher-Depth} 
& \multicolumn{2}{c}{\rule{0pt}{2.1ex}\textbf{TIDY}-Denoised} \\ 

\rule{0pt}{2.1ex}
& AbsRel $\downarrow$ & \phantom{0}$\delta_1$ $\uparrow$\phantom{0}
& AbsRel $\downarrow$ & \phantom{0}$\delta_1$ $\uparrow$\phantom{0}
& AbsRel $\downarrow$ & \phantom{0}$\delta_1$ $\uparrow$\phantom{0} \\ 
\hline

indoor-dark-aggresive   & 0.264 & 0.609 & 0.603 & 0.014& \textbf{0.257} & \textbf{0.610} \\
indoor-global-aggresive & 0.241 & 0.691 & 0.599 & 0.016 & \textbf{0.217} & \textbf{0.711} \\
indoor-local-aggresive  & 0.218 & 0.695 & 0.603 & 0.013 & \textbf{0.196} & \textbf{0.725} \\
\hline \hline
\end{tabular}%
}
\vspace{-5mm}
\end{table}

\begin{table}[b]
\vspace{-4mm}
\caption{Ablation Study of Methods on IRE and SCaN-TIR Dataset. Best results are highlighted in \textbf{bold} and second best are \underline{underlined}.}
\label{tab:ablation1}
\centering
\setlength{\tabcolsep}{3pt}
\resizebox{\columnwidth}{!}{%
\begin{tabular}{M{0.7cm} M{0.7cm} M{0.7cm} M{0.72cm}|cc|cc|c}
\hline \hline
\multicolumn{4}{c|}{\rule{0pt}{2.5ex}Method} 
& \multicolumn{2}{c|}{\rule{0pt}{2.5ex}IRE Dataset} 
& \multicolumn{2}{c|}{\rule{0pt}{2.5ex}SCaN-TIR Dataset} 
& \multirow{2}{*}{GFLOPS $\downarrow$} \\

\rule{0pt}{2.5ex}DWT & \rule{0pt}{2.5ex}FiLM 
& \rule{0pt}{2.5ex}\scalebox{0.9}{$\mathcal{L}_{\mathrm{WE}}$} 
& \rule{0pt}{2.5ex}\scalebox{0.75}{$\mathcal{L}_{\mathrm{WDSI}}$}
& \rule{0pt}{2.5ex}PSNR $\uparrow$ 
& \rule{0pt}{2.5ex}SSIM $\uparrow$ 
& \rule{0pt}{2.5ex}PSNR $\uparrow$ 
& \rule{0pt}{2.5ex}SSIM $\uparrow$ & \\ 

\hline

 & & & 
& 35.67 & 0.9496 
& 13.39 & 0.351 
& 509.06 \\

\checkmark & & & 
& 35.68 & 0.9386 
& 13.35 & 0.342 
& \textbf{128.64} \\

\checkmark & \checkmark & & 
& 36.25 & 0.9503 
& 13.56 & 0.357 
& \underline{129.66} \\

\checkmark & \checkmark & \checkmark & 
& \textbf{36.51} & \textbf{0.9546} 
& 13.58 & 0.355 
& \underline{129.66} \\

\checkmark & \checkmark &  & \checkmark 
& 36.26 & 0.9501 
& \underline{13.83} & \underline{0.357} 
& \underline{129.66} \\

\checkmark & \checkmark & \checkmark & \checkmark 
& \underline{36.44} & \underline{0.9519} 
& \textbf{14.11} & \textbf{0.358} 
& \underline{129.66} \\

\hline \hline
\end{tabular}%
}
\end{table}

\subsection{Ablation Study}

We perform comprehensive ablation studies to validate key components of our proposed framework, evaluating the impact of \ac{DWT} encoder/decoder (DWT Enc.) framework, FiLM, and the two novel proposed loss functions. \tabref{tab:ablation1} shows the effectiveness of these components on the external dataset, indicating that the implementation of DWT, FiLM and $\mathcal{L}_\mathrm{WE}$ shows the best results on IRE dataset composed of only artificial stochastic noise, and adding the real \ac{FPN} specific loss $\mathcal{L}_\mathrm{WDSI}$ for SCaN-TIR dataset shows significant improvement, all while notably reducing the computational cost by approximately a quarter of the original GFLOPS, similar to the theoretical decrease mentioned in \secref{sec:dwt_film}.

\begin{figure}[t]

    \centering
    \includegraphics[width=0.95\columnwidth]{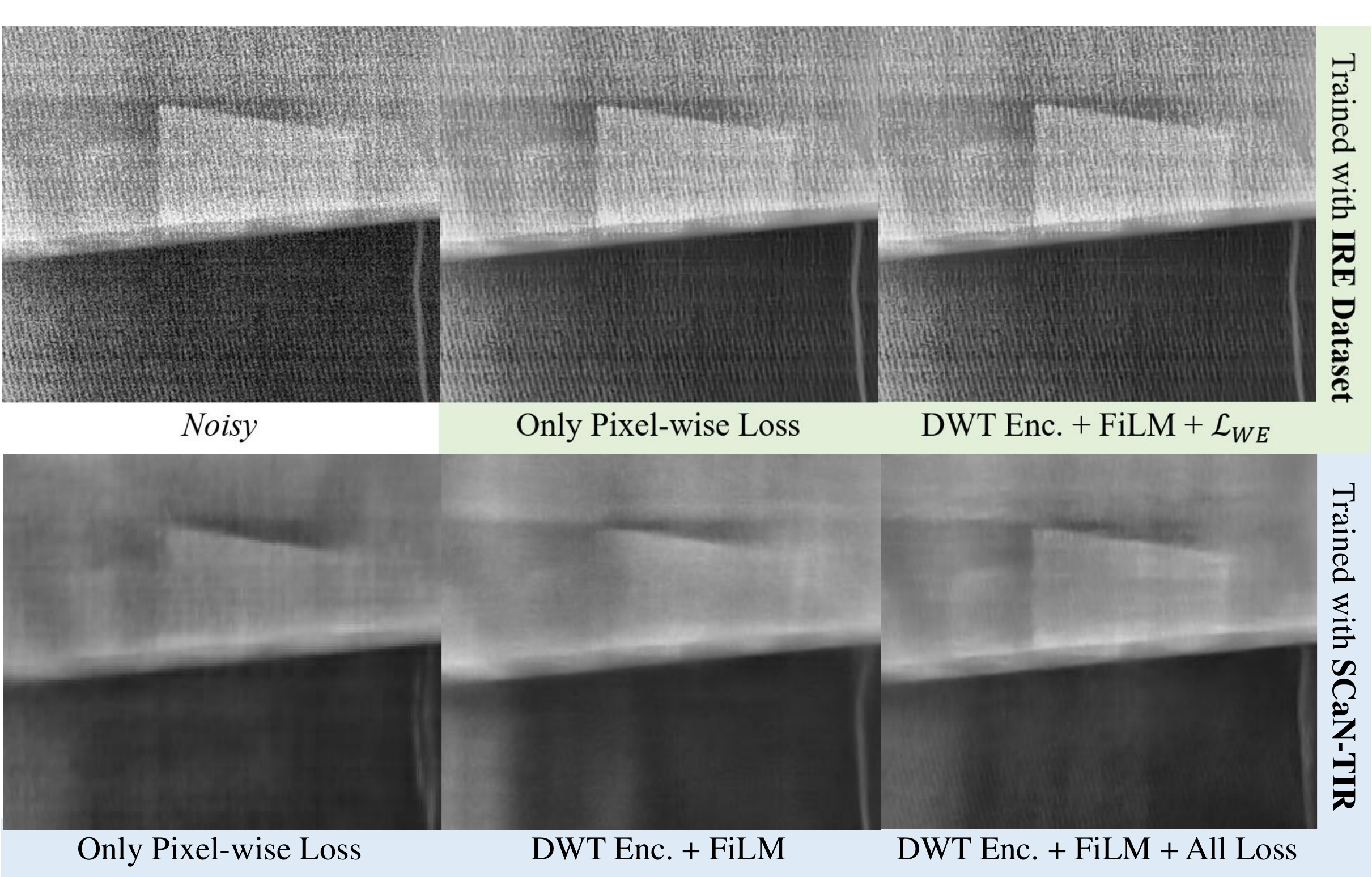}
    \caption{\textbf{Qualitative Ablation Study} on zero-shot results of Multi-Spectral dataset image. \textbf{First Row:} Models trained on IRE with only pixel-wise loss show minimal denoising due to the synthetic-real domain gap. \textbf{Second Row:} Adding DWT Enc. and FiLM improves suppression and structure preservation, with the proposed losses further enhancing clarity and reducing stripe artifacts.}
    \label{fig:ablation}
    \vspace{-6mm}
\end{figure}

Furthermore, \figref{fig:ablation} shows qualitative ablation studies via zero-shot denoising results on Multi-Spectral dataset image. The first row presents results from the model trained on the IRE dataset, exhibiting minimal denoising efficacy due to the domain gap between synthetic and real-world noise characteristics. The second row shows results from TIDY trained on the SCaN-TIR dataset, demonstrating substantial denoising performance. The incorporation of DWT Enc. and FiLM yields stronger noise suppression, albeit with slightly softened image details. When all proposed loss functions are applied, the resulting output exhibits clearer denoising, suggesting the effectiveness of both the DWT encoder and FiLM in denoising performance accompanied by significant decrease in model complexity, and the capability of the novel losses to enhance structural preservation by distinguishing meaningful features from noise. 

\section{Future Works and Conclusion}
\label{sec:conclusion}



TIDY offers a lightweight solution for noisy \ac{TIR} imaging in downstream robotic tasks, originally solved via heavy and complex methods. 
While TIDY shows robust denoising performance and online execution, several limitations remain. As with any data-driven model, performance could be further enhanced by access to larger volumes of real paired training data. Although TIDY already achieves robust per-frame denoising, incorporating sequential inputs and temporal consistency offer a promising direction to further improve stability and accuracy in video-based applications. 


\balance
\small
\bibliographystyle{IEEEtranN} 
\bibliography{string-short,references}

@book{pathak2009the_wavelet_transform,
  title={The wavelet transform},
  author={Pathak, Ram Shankar},
  volume={4},
  year={2009},
  publisher={Springer Science \& Business Media}
}

@article{haar1911theorie,
  title={Zur theorie der orthogonalen funktionensysteme},
  author={Haar, Alfred},
  journal={Mathematische Annalen},
  volume={71},
  number={1},
  pages={38--53},
  year={1911},
  publisher={Springer}
}

@article{skodras2003discrete_dwtintroduction,
  title={Discrete wavelet transform: an introduction},
  author={Skodras, Athanassios N},
  journal=T-HUTC,
  volume={2},
  number={1},
  pages={1--26},
  year={2003}
}

@article{liu2025twostage_irenhance,
  title={A two-stage approach for single thermal image restoration},
  author={Liu, Guanyu and Xu, Jinxiang and Cheng, Yihui and Su, Yi and Yang, Biwen},
  journal=J-EL,
  volume={61},
  number={1},
  pages={e70111},
  year={2025},
  publisher={Wiley Online Library}
}

@article{ding2023u2d2net,
  title={U2D2Net: Unsupervised unified image dehazing and denoising network for single hazy image enhancement},
  author={Ding, Bosheng and Zhang, Ruiheng and Xu, Lixin and Liu, Guanyu and Yang, Shuo and Liu, Yumeng and Zhang, Qi},
  journal=IEEE_J_MM,
  volume={26},
  pages={202--217},
  year={2023},
  publisher={IEEE}
}

@article{zhang2008multiresolution_bilateral_filtering,
  title={Multiresolution bilateral filtering for image denoising},
  author={Zhang, Ming and Gunturk, Bahadir K},
  journal=IEEE_J_IP,
  volume={17},
  number={12},
  pages={2324--2333},
  year={2008},
  publisher={IEEE}
}

@article{he2018single_image_based_nonuniformity-snrcnn,
  title={Single-image-based nonuniformity correction of uncooled long-wave infrared detectors: A deep-learning approach},
  author={He, Zewei and Cao, Yanpeng and Dong, Yafei and Yang, Jiangxin and Cao, Yanlong and Tisse, Christel-L{\"o}ic},
  journal=J-AP,
  volume={57},
  number={18},
  pages={D155--D164},
  year={2018},
  publisher={OSA}
}

@inproceedings{saragadam2021deepir,
  title={Thermal image processing via physics-inspired deep networks},
  author={Saragadam, Vishwanath and Dave, Akshat and Veeraraghavan, Ashok and Baraniuk, Richard G},
  booktitle=C-CVPR,
  pages={4057--4065},
  year={2021}
}

@inproceedings{li2022odombeyondvision,
  title={Odombeyondvision: An indoor multi-modal multi-platform odometry dataset beyond the visible spectrum},
  author={Li, Peize and Cai, Kaiwen and Saputra, Muhamad Risqi U and Dai, Zhuangzhuang and Lu, Chris Xiaoxuan},
  booktitle=C-IROS,
  pages={3845--3850},
  year={2022}
}

@article{gil2024fieldscale,
  title={Fieldscale: Locality-aware field-based adaptive rescaling for thermal infrared image},
  author={Gil, Hyeonjae and Jeon, Myung-Hwan and Kim, Ayoung},
  journal=IEEE_J_RAL,
  year={2024},
  publisher={IEEE}
}

@inproceedings{dai2021multispectraldataset,
  title={A multi-spectral dataset for evaluating motion estimation systems},
  author={Dai, Weichen and Zhang, Yu and Chen, Shenzhou and Sun, Donglei and Kong, Da},
  booktitle=C-ICRA,
  pages={5560--5566},
  year={2021}
}

@inproceedings{liu2025deal,
  title={DEAL: Data-Efficient Adversarial Learning for High-Quality Infrared Imaging},
  author={Liu, Zhu and Wang, Zijun and Liu, Jinyuan and Meng, Fanqi and Ma, Long and Liu, Risheng},
  booktitle=C-CVPR,
  pages={28198--28207},
  year={2025}
}

@article{liu2023thermal_tv-dip,
  title={Thermal imaging spatial noise removal via deep image prior and step-variable total variation regularization},
  author={Liu, Kang and Chen, Honglei and Bao, Wenzhong and Wang, Jianlu},
  journal=J-IPT,
  volume={134},
  pages={104888},
  year={2023},
  publisher={Elsevier}
}

@inproceedings{carmichael2025trnerf,
  title={TRNeRF: Restoring Blurry, Rolling Shutter, and Noisy Thermal Images with Neural Radiance Fields},
  author={Carmichael, Spencer and Bhat, Manohar and Ramanagopal, Mani and Buchan, Austin and Vasudevan, Ram and Skinner, Katherine A},
  booktitle=C-WACV,
  pages={7980--7990},
  year={2025}
}

@article{yang2024destripecyclegan,
  title={DestripeCycleGAN: Stripe simulation CycleGAN for unsupervised infrared image destriping},
  author={Yang, Shiqi and Qin, Hanlin and Yuan, Shuai and Yan, Xiang and Rahmani, Hossein},
  journal=IEEE_J_IM,
  year={2024},
  publisher={IEEE}
}

@article{guan2019wavelet_dnn_stripe,
  title={Wavelet deep neural network for stripe noise removal},
  author={Guan, Juntao and Lai, Rui and Xiong, Ai},
  journal=IEEE_O_ACC,
  volume={7},
  pages={44544--44554},
  year={2019},
  publisher={IEEE}
}

@inproceedings{chen2022nafnet,
  title={Simple baselines for image restoration},
  author={Chen, Liangyu and Chu, Xiaojie and Zhang, Xiangyu and Sun, Jian},
  booktitle=C-ECCV,
  pages={17--33},
  year={2022},
  organization={Springer}
}

@inproceedings{perez2018film,
  title={FiLM: Visual reasoning with a general conditioning layer},
  author={Perez, Ethan and Strub, Florian and De Vries, Harm and Dumoulin, Vincent and Courville, Aaron},
  booktitle=C-AAAI,
  volume={32},
  number={1},
  year={2018}
}

@article{qin2018vinsmono,
  title={Vins-mono: A robust and versatile monocular visual-inertial state estimator},
  author={Qin, Tong and Li, Peiliang and Shen, Shaojie},
  journal=IEEE_J_RO,
  volume={34},
  number={4},
  pages={1004--1020},
  year={2018},
  publisher={IEEE}
}

@article{saputra2020deeptio,
  title={Deeptio: A deep thermal-inertial odometry with visual hallucination},
  author={Saputra, Muhamad Risqi U and others},
  journal=IEEE_J_RAL,
  volume={5},
  number={2},
  pages={1672--1679},
  year={2020},
  publisher={IEEE}
}

@article{thrhee-2025-icra-ws,
    AUTHOR = {Rhee, Tai Hyoung and Lee, Dong-Guw and Kim, Ayoung},
    TITLE = {TIR-Diffusion: Diffusion-based Thermal Infrared Image Denoising via Latent and Wavelet Domain Optimization},
    journal = {in Proc. IEEE Intl. Conf. Robot. Autom. Workshops},
    YEAR = {2025},
    ADDRESS = {Atlanta, GA, USA},
}

@article{carmichael2025nsavp,
  title={Dataset and benchmark: Novel sensors for autonomous vehicle perception},
  author={Carmichael, Spencer and Buchan, Austin and Ramanagopal, Mani and Ravi, Radhika and Vasudevan, Ram and Skinner, Katherine A},
  journal=J-IJRR,
  volume={44},
  number={3},
  pages={355--365},
  year={2025},
  publisher={SAGE Publications Sage UK: London, England}
}

@article{zuo2025monotherdepth,
  title={MonoTher-Depth: Enhancing Thermal Depth Estimation via Confidence-Aware Distillation},
  author={Zuo, Xingxing and Ranganathan, Nikhil and Lee, Connor and Gkioxari, Georgia and Chung, Soon-Jo},
  journal=IEEE_J_RAL,
  year={2025},
  publisher={IEEE}
}

@article{shin2021self,
  title={Self-supervised depth and ego-motion estimation for monocular thermal video using multi-spectral consistency loss},
  author={Shin, Ukcheol and Lee, Kyunghyun and Lee, Seokju and Kweon, In So},
  journal=IEEE_J_RAL,
  volume={7},
  number={2},
  pages={1103--1110},
  year={2021},
  publisher={IEEE}
}

@inproceedings{shin2023ms2,
  title={Deep depth estimation from thermal image},
  author={Shin, Ukcheol and Park, Jinsun and Kweon, In So},
  booktitle=C-CVPR,
  pages={1043--1053},
  year={2023}
}

@article{lee2024thermalchameleon,
  title={Thermal Chameleon: Task-Adaptive Tone-Mapping for Radiometric Thermal-Infrared Images},
  author={Lee, Dong-Guw and Kim, Jeongyun and Cho, Younggun and Kim, Ayoung},
  journal=IEEE_J_RAL,
  year={2024},
  publisher={IEEE}
}

@inproceedings{song2024discocal,
  title={Unbiased estimator for distorted conics in camera calibration},
  author={Song, Chaehyeon and Shin, Jaeho and Jeon, Myung-Hwan and Lim, Jongwoo and Kim, Ayoung},
  booktitle=C-CVPR,
  pages={373--381},
  year={2024}
}

@inproceedings{liu2025hmtir,
  title={Enhancing Infrared Vision: Progressive Prompt Fusion Network and Benchmark},
  author={Liu, Jinyuan and Chen, Zihang and Liu, Zhu and Jiang, Zhiying and Ma, Long and Fan, Xin and Liu, Risheng},
  booktitle=C-NIPS,
  year={2025}
}

@article{lee2022vivid++,
  title={ViViD++: Vision for visibility dataset},
  author={Lee, Alex Junho and Cho, Younggun and Shin, Young-sik and Kim, Ayoung and Myung, Hyun},
  journal=IEEE_J_RAL,
  volume={7},
  number={3},
  pages={6282--6289},
  year={2022},
  publisher={IEEE}
}

@article{lin2025depthanything3,
  title={Depth anything 3: Recovering the visual space from any views},
  author={Lin, Haotong and Chen, Sili and Liew, Junhao and Chen, Donny Y and Li, Zhenyu and Shi, Guang and Feng, Jiashi and Kang, Bingyi},
  journal={arXiv preprint arXiv:2511.10647},
  year={2025}
}

@article{cruz2021autonomous,
  title={Autonomous thermal vision robotic system for victims recognition in search and rescue missions},
  author={Cruz Ulloa, Christyan and Prieto S{\'a}nchez, Guillermo and Barrientos, Antonio and Del Cerro, Jaime},
  journal={Sensors},
  volume={21},
  number={21},
  pages={7346},
  year={2021},
  publisher={MDPI}
}

@inproceedings{wong2009effective,
  title={An effective surveillance system using thermal camera},
  author={Wong, Wai Kit and Tan, Poi Ngee and Loo, Chu Kiong and Lim, Way Soong},
  booktitle={IEEE Intl. Conf. Signal Acquis. Process.},
  pages={13--17},
  year={2009},
}

@inproceedings{zhao2020tptio,
  title={Tp-tio: A robust thermal-inertial odometry with deep thermalpoint},
  author={Zhao, Shibo and Wang, Peng and Zhang, Hengrui and Fang, Zheng and Scherer, Sebastian},
  booktitle=C-IROS,
  pages={4505--4512},
  year={2020},
}

@article{shin2022maximizing,
  title={Maximizing self-supervision from thermal image for effective self-supervised learning of depth and ego-motion},
  author={Shin, Ukcheol and Lee, Kyunghyun and Lee, Byeong-Uk and Kweon, In So},
  journal=IEEE_J_RAL,
  volume={7},
  number={3},
  pages={7771--7778},
  year={2022}
}

@inproceedings{lee2022sequential,
  title={Sequential thermal image-based adult and baby detection robust to thermal residual heat marks},
  author={Lee, Dong-Guw and Song, Kyu-Seob and Nho, Young-Hoon and Kim, Ayoung and Kwon, Dong-Soo},
  booktitle=C-IROS,
  pages={13120--13127},
  year={2022}
}

@STRING{C-AAAI = {Proc. {AAAI} National Conf. on Art. Intell.}}

@STRING{C-CVPR = {Proc. {IEEE} Conf. on Comput. Vision and Pattern Recog.}}

@STRING{C-ECCV = {Proc. European Conf. on Comput. Vision}}

@STRING{C-ICRA = {Proc. {IEEE} Intl. Conf. on Robot. and Automat.}}

@STRING{C-IROS = {Proc. {IEEE}/{RSJ} Intl. Conf. on Intell. Robots and Sys.}}

@STRING{C-NIPS = {Adv. Neural Inf. Process. Syst.}}

@STRING{IEEE_J_IP = {{IEEE} Trans. Image Processing}}

@STRING{IEEE_J_MM = {{IEEE} Trans. Multimedia}}

@STRING{IEEE_J_RAL = {{IEEE} Robot. and Automat. Lett.}}

@STRING{IEEE_J_RO = {{IEEE} Trans. Robot.}}

@STRING{IEEE_J_IM = {{IEEE} Trans. Instrum. and Meas.}}

@STRING{IEEE_O_ACC = {{IEEE} Access}}

@STRING{J-IJRR = {Intl. J. of Robot. Research}}

@STRING{T-HUTC = {Tech. Rep., Hellenic Open Univ.}}

@STRING{J-EL = {Electron. Lett.}}

@STRING{J-AP = {Appl. Opt.}}

@STRING{J-IPT = {Infrared Phys. Tech.}}

@STRING{C-WACV = {Proc. {IEEE} Winter Conf. Appl. Comput. Vis.}}

\end{document}